# CBAM Integrated Attention Driven Model For Betel Leaf Diseases Classification With Explainable AI

Sumaiya Tabassum
*Department of Computer Science and Engineering*
*Dhaka International University*
Dhaka, Bangladesh
sumaiyatabassum230@gmail.com

Md. Faysal Ahamed
*Department of Electrical and Computer Engineering*
*Rajshahi University of Engineering and Technology*
Rajshahi, Bangladesh
faysal.ahamed@ece.ruet.ac.bd

*Abstract*—Betel leaf is an important crop because of its economic advantages and widespread use. Its betel vines are susceptible to a number of illnesses that are commonly referred to as betel leaf disease. Plant diseases are the largest threat to the food supply's security, and they are challenging to identify in time to stop possible financial damage. Interestingly, artificial intelligence can leave a big mark on the betel leaf industry since it helps with output growth by forecasting sickness. This paper presents a lightweight CBAM-CNN model with just 2.13 million parameters (8.13 MB), incorporating CBAM (Convolutional Block Attention Module) to improve feature emphasis without depending on heavy pre-trained networks. The model's capacity to discern minute variations among leaf disease classes is improved by the integrated attention mechanism, which allows it to adaptively focus on significant spatial and channel-wise information. In order to ensure class balance and diversity for efficient model training and validation, this work makes use of an enriched dataset of 10,185 images divided into three categories: Healthy Leaf, Leaf Rot, and Leaf Spot. The proposed model achieved a precision of 97%, recall of 94%, and F1 score of 95%, and 95.58% accuracy on the test set demonstrating strong and balanced classification performance outperforming traditional pre trained CNN models. The model's focus regions were visualized and interpreted using Grad-CAM (Gradient-weighted Class Activation Mapping), an explainable AI technique.

## I. Introduction

In tropical Southeast Asian nations, including India, Bangladesh, Malaysia, Pakistan, and Sri Lanka, betel leaf (Piper betle), sometimes referred to locally as paan, is a cash crop with cultural and economic significance[1]. Generation after generation has continued to demand it due to its close ties to literature, tradition, and social rituals. Betel leaf cultivation has grown significantly in Bangladesh in recent years. The Kushtia district collected betel leaves on 2,230 hectares of land in 2021–2022, a 4% increase from the year before, according to the DAE[2]. More recently, in the first seven months of the fiscal year 2023–2024, and from 4,580 hectares, the Rajshahi district generated over 77,219 tonnes of betel leaves, generating about Tk 25 billion for the local economy[3]. More than 39,000 betel producers are registered in this district alone, and the crop's profits have eclipsed those of more conventional fruits like mangoes. In 2024, more than 4,000 smallholder farmers in southern areas like Barguna grew betel leaves over 385 hectares; under ideal circumstances, some of them reported profits of up to Tk 200,000 per bigha[4]. These changes are in line with a nationwide trend of increasing betel leaf output, which is bolstered by the plant's perennial nature, potential for export, and robust domestic demand. Consequently, the betel leaf sector remains a crucial element of rural livelihoods and agroeconomic growth throughout South Asia.

Originating from the Piperaceae family, betel leaf [5] is a traditional medicinal herb that is widely used in Asian countries. The betel vine's antimicrobial and antioxidant qualities make it useful for preservation. It also has anticancer properties and protection against fungal infections. It keeps the human body's blood glucose levels stable and speeds up wound healing. It increases metabolism and prevents stomach ulcer disorders. Additionally, betel leaves are regarded as a beneficial herb that contains bioactive phenolic compounds, and the plant's extracts are valuable in a variety of industries, including organic synthesis, food and beverage, and medicine. The manufacture of betel leaves surely helps rural farmers' livelihoods by including them in gathering and wrapping. However, the true problem arises when the intake are afflicted with various illnesses that compromise their freshness, such as bacterial leaf disease, fungal brown spots, and dried leaf disease. Bacteria like Xanthomonas campestris pv. betlicola produces bacterial leaf disease, whereas fungus like Cercospora beticola and Colletotrichum capsici cause fungal brown spot disease[6].

Prior studies on the categorization of betel leaves have mostly used deep learning and classical machine learning techniques, which frequently call for substantial manual feature engineering or large labelled datasets. Numerous studies used complex architectures with millions of parameters, such as deep CNNs or ViT, which are inappropriate for use in agricultural environments with limited resources. Furthermore, explainable AI approaches have received little attention, which has resulted in a gap in decision interpretability and model transparency. Farmers and agricultural specialists are examples of end users whose trust and practical adoption are hampered by this lack of explainability. In this work, we present a CBAM-integrated convolutional neural network (CBAM-CNN) specifically designed for the classification of the health state of betel leaves. In order to capture both global and local features, the design uses hierarchical convolution blocks with descending kernel sizes (7→5→3).CBAM is used to improve each block in order to highlight pertinent spatial and channel-wise information. Better generalization in the categorization of leaf diseases is supported by this approach, which makes feature

learning effective even with little data. The contributions of this work are:

1) We integrated CBAM modules after every block to improve feature extraction by adaptively fine-tuning spatial and channel-wise attention. This approach incorporates CBAM to improve focus on crucial spatial and channel information and uses gradually decreasing kernel sizes to capture multiscale texture features.
2) We suggested a model providing a lightweight solution that is perfect for deployment in resource-constrained or edge situations without sacrificing accuracy because of its compact architecture, which has only 2.13 million parameters (~8.13 MB).
3) We incorporated Grad-CAM visualization to see which areas of the betel leaf images the model concentrates on during categorization. In sensitive agricultural and medical imaging applications, in particular, this approach enhances model transparency and fosters confidence in the decision-making process.

Section II of the study reviews the literature, Section III is about the information of dataset, and Section IV covers the research approach, which includes the suggested model architecture, Section V provides Results and Discussion, and Section VI covers the comparison with previous works, Section VII presents Grad CAM Visualization, and Section VII covers the limitations and Section IX covers the Conclusion.

## II. LITERATURE REVIEW

As a new area of study, accurate identification of betel leaf disease is essential to enable thorough ethical trading practices. Thus, several studies evaluated ML and DL models.S. Kusuma and K. R. Jothi et al. developed a study that examines how well the deep learning models,such as vision transformer (ViT), DenseNet201, ResNet152V2, and VGG19 detect illnesses in betel leaves through image analysis[7]. With the best accuracy (98.77%) among them, DenseNet201 demonstrated its promise for early plant disease diagnosis to promote food security and lower crop loss. M. Tahsin et al. presented an informative framework for betel leaf disease detection using DenseNet-201, which achieved 99.23% accuracy[8]. It further integrates semi-supervised learning (FixMatch, MixMatch, MeanTeacher) and explainable AI (XAI) techniques to enhance classification with minimal labeled data, enabling real-time, transparent, and efficient disease identification. The framework's flexibility across real-world scenarios may be limited by its reliance on a single deep learning backbone (DenseNet-201). R.H. Hridoy et al. suggested a deep learning-based technique using a dataset of 10,662 photos for the quick and precise recognition of betel plant diseases. With the greatest accuracy of 98.84% among the tested models, EfficientNet B5 proved to be a robust model with low misclassification, making it useful for managing illness and guaranteeing quality in the betel leaf sector[9].M. A. Malek et al. used both conventional machine learning methods and deep learning models, such as VGG16, VGG19, ResNet50, AlexNet, and InceptionV3, to investigate the detection of betel leaf illnesses. With the maximum accuracy of 94.83% on a dataset of 5,800 preprocessed photos, InceptionV3 proved its usefulness in identifying betel leaf illnesses photographed under actual conditions[10].The stated maximum accuracy of 94.83%, however, indicates that there is still opportunity for improvement in managing intricate variations of betel leaf illnesses, even after testing several models. F. David et al. presented a study in which betel leaf images are segmented using a region of interest (ROI) technique. Features are extracted using GLCM, and the Extreme Learning Machine (ELM) algorithm is used for classification[11].The technique demonstrated ELM's promise for quick and accurate diagnosis of plant leaf diseases in agricultural applications by achieving 97% accuracy using a dataset of 1,047 pictures spanning five disease classes. Nevertheless, the study was constrained by a somewhat small collection of 1,047 photos, which would limit how far the 97% accuracy observed can be used. Using colour features, H. Hamdani et al. proposed an image processing-based technique to categorize Piper betle L. leaves into red, green, and black variants. They used Naïve Bayes, SVM, and k-NN classifiers for classification, ROI detection, and feature extraction[12].The study demonstrated the excellent discriminative ability of colour characteristics for betel leaf identification by reporting a 100% accuracy rate using the k-NN classifier. The strength of the categorization approach is limited by the study's exclusive emphasis on coloured data, which ignores other potentially significant aspects. A dataset of 4,156 high-resolution betel leaf photos covering 15 diseases was created by M.Gayakwad et al. and arranged hierarchically by category[13]. They reported classification accuracies between 0.7 and 0.9 using Vision Transformer. The model may have trouble with some illness classes, though, as seen by the reported accuracy range (0.7–0.9), which calls for more optimization or larger, balanced datasets. Many of the studies carried out so far have mostly relied on pre-trained deep learning models like DenseNet201, ResNet152V2, and EfficientNetB5.These models have large model sizes and millions of parameters, which makes them computationally costly and less appropriate for deployment in environments with limited resources. Furthermore, explainability is frequently neglected or only partially addressed, which compromises interpretability and confidence in practical implementations. In order to improve feature focus while preserving a small architecture, this proposed research work suggests a lightweight CBAM-CNN model that incorporates CBAM. In addition to lowering the number of parameters, the model uses explainable AI techniques based on Grad-CAM to guarantee decision-making transparency. By emphasizing the areas of an image that have the greatest influence on a model's prediction, Grad-CAM offers visually understandable explanations.

## III. DATASET DESCRIPTION

The dataset used in this study contains 12,222 JPEG pictures of betel leaves with a resolution of 1024 x 1024 pixels[14].Over the course of two years (2020–2022), 2,037 original images were gathered from four distinct betel growing areas in Mymensingh, Bangladesh, and 10,185 augmented images were produced utilizing methods such as flipping, brightness correction, contrast variation, and rotation. To guarantee quality and relevance, the photos were taken in the daylight under the guidance of plant pathology specialists and are divided into three classes: Healthy Leaf, Leaf Rot, and Leaf Spot. This study uses the augmented dataset to validate the methodology. TABLE I shows the classwise distribution of betel leaf diseases from the dataset. Additionally, Fig. 1 displays the class-wise samples from the dataset of the three classes.

TABLE I. PORTRAYAL OF THE DATASET

| Class Name | No of samples (Original Dataset) | No of samples (Augmented Dataset) |
|---|---|---|
| Healthy Leaf | 1080 | 5400 |
| Leaf Rot | 269 | 1345 |
| Leaf Spot | 688 | 3440 |
| **Total images** | **2037** | **10,185** |

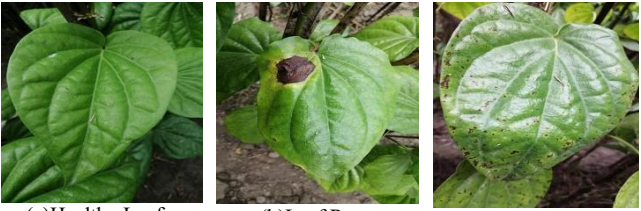

(a)Healthy Leaf     (b)Leaf Rot     (c) Leaf Spot

Fig. 1. The dataset includes three types of leaf (a) Healthy Leaf (b) Leaf Rot (c) Leaf Spot

## IV. RESEARCH METHODOLOGY

### A. Dataset Pre Processing

To guarantee compatibility with CNN-based architectures, the augmented dataset was first enlarged to 224×224 with a three-channel RGB format before being arranged into class-specific folders. One aspect of preprocessing was utilizing directory-based class labelling to organize the data for image generators. To maintain class distribution, stratified sampling was then used to divide the total dataset into training (80%), validation (10%), and test (10%) sets. Model training and validation were accomplished using the training and validation sets, respectively. The test set was used to assess the model's performance on unseen data. This separation ensures an objective assessment of the model's generalization capabilities by guaranteeing that its performance is evaluated on novel data.

### B. CBAM Integrated Attention Based Customized CNN Architecture

Fig. 2 shows the architecture of the proposed customized and attention-mechanisms integrated model, which consists of four attention module integrated blocks, each with two convolutional layers. A CBAM attention module and max pooling finish the first block, which has two Conv2D layers with 7×7 kernels and 32 filters each, followed by Batch Normalization and ReLU. This large kernel aids in the input image's global spatial feature extraction. Two Conv2D layers with 5x5 kernel and 64 filters are included in the second block, which also includes BatchNormalization and ReLU after each convolution layer. CBAM and pooling are used to wrap up. The model's third and fourth blocks both use comparable structures. Two Conv2D layers with 3×3 kernels and 128 filters make up the third block, which is designed to extract local fine-grained features from mid-level representations. In a similar manner, the fourth block has two Conv2D layers with 256 filters and 3×3 kernels that aid in identifying more intricate localized patterns in the input. To stabilize and improve learning, Batch Normalization and ReLU activation come after each of these Conv2D layers. Additionally, a CBAM attention module is integrated into both blocks to improve the feature maps by highlighting descriptive regions and repressing less valuable ones. Max pooling is included after each block too. After two fully linked layers (1024 and 512 units) with dropout and Global Average Pooling, the network culminates in a softmax classification layer. Since the CBAM module adaptively refines the learnt features by highlighting the most pertinent channels and spatial regions at each stage, it is essential to integrate it after each convolutional block. The model's ability to discriminate and generalize is enhanced by this progressive attention mechanism, which helps it concentrate more efficiently on significant patterns across the network.It refines valuable features and thus improves the model's ability to recognize different classes and thus achieves great performance.

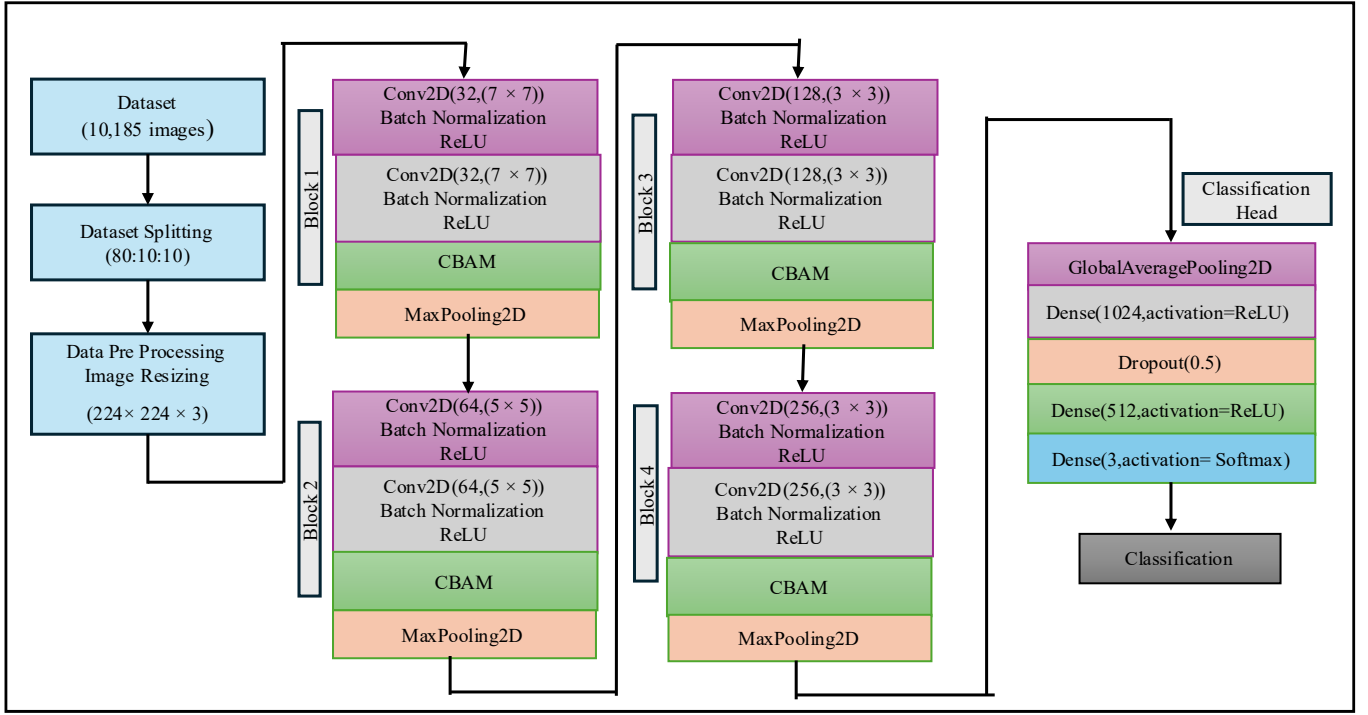

Fig. 2. The proposed model architecture

## C. Attention Mechanisms

The attention module ignores the unimportant areas of the image and emphasizes the important aspects.

CBAM(Convolutional Block Attention Module):

CBAM is a general-purpose, lightweight attention module that works with any Convolutional Neural Network (CNN) design introduced by Sanghyun Woo et al.[15].By emphasizing significant elements along the spatial and channel dimensions, it improves the network's representational power. Channel Attention (CA) and Spatial Attention (SA) are its two sub-modules, shown in Fig. 3.

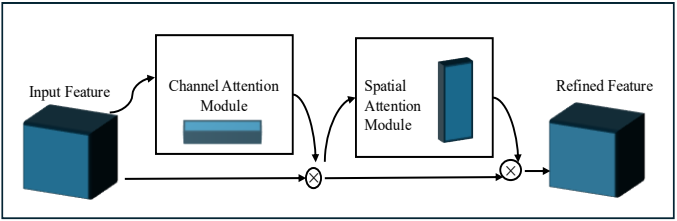

Fig. 3. Convolutional block attention module

**Channel Attention (CA):** Each feature channel is given a different weight by the CA module, which aggregates global data via average and max pooling. This approach enhances feature discrimination and ultimately contributes to the better results shown by helping the network concentrate on the most information channels (such as colour or texture patterns crucial for identifying leaf diseases). The definition of the attention mechanism is:

$$M_c(F) = \sigma(MLP(AvgPool(F)) + MLP(MaxPool(F)))\ldots(1)$$

Where: $F \in \mathbb{R}^{C \times H \times W}$ is the input feature map, AvgPool and MaxPool apply global average and max pooling over the spatial dimensions, $MLP$ is a shared two-layer multilayer perceptron with ReLU activation, $\sigma$ is the sigmoid activation function, $M_c(F) \in \mathbb{R}^{C \times 1 \times 1}$ is the channel attention map. The refined feature map is computed by element-wise multiplication:

$$F' = M_c(F) \odot F \ldots\ldots\ldots\ldots(2)$$

**Spatial Attention (SA):** After CA, the SA Module focuses on identifying important regions within the feature map. It compresses the input along the channel axis using average and max operations, followed by a convolution:

$$M_s(F') = \sigma\big(f^{7\times 7}([AvgPool(F'); MaxPool(F')])\big)\ldots(3)$$

Where:[;] denotes channel-wise concatenation, $f^{7\times 7}$ is a convolutional layer with a $7 \times 7$ kernel, $M_s(F') \in \mathbb{R}^{1 \times H \times W}$ is the spatial attention map. Finally, the spatially refined feature map is:

$$F'' = M_s(F') \odot F' \ldots\ldots\ldots\ldots(4)$$

## D. Experimental Set Up

This study used a GPU P100 to run our full experiment on Kaggle. The proposed model was trained over 300 epochs using a batch size of 64 images per epoch . With a learning rate of 0.001, we used the Adam optimizer, and the categorical cross-entropy loss function guided the optimization process. The hyperparameters are described in TABLE II.

## V. RESULTS AND DISCUSSION

This section gives empirical and graphical findings to evaluate the classification performance, with a major focus on examining the effectiveness of the attention module in improving output.

TABLE II. HYPERPARAMETER SETTING FOR THE PROPOSED STUDY

| Parameter Name | Attribute |
|---|---|
| Weights | Custom (random initialization) |
| Verbose | 2 |
| Epochs | 300 |
| Metrics | Accuracy |
| Batch Size | 64 |
| Loss Function | Categorical Crossentropy |
| Optimizer | Adam |
| Learning rate | 0.001 |

The confusion matrix for the test set using our suggested model is displayed in Fig. 4. 539 out of 540 samples were accurately classified by the model, which did best on Healthy_Leaf. With 124 accurate predictions and a few misclassifications, Leaf_Rot was likewise classified well. The Leaf_Spot class had the most mistakes, with 31 samples incorrectly predicted as Healthy_Leaf. The model performs well overall, with only a slight confusion between Leaf_Spot and Healthy_Leaf.

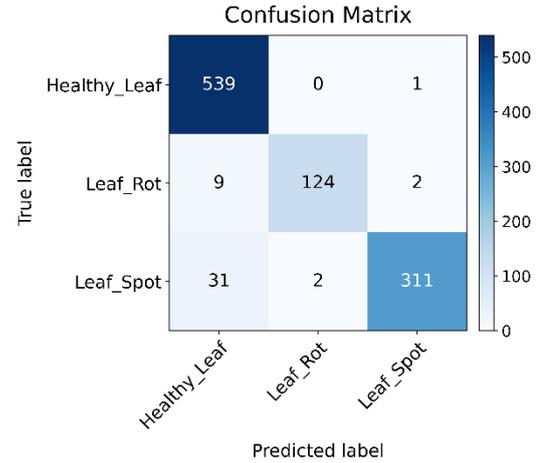

Fig. 4. Confusion matrix of the proposed model

TABLE III. CLASSWISE PERFORMANCE OF THE PROPOSED MODEL

| Class Name | Precision(0-1) | Recall(0-1) | F1 Score (0-1) |
|---|---|---|---|
| Healthy Leaf(0) | 0.93 | 1.00 | 0.96 |
| Leaf Rot(1) | 0.98 | 0.92 | 0.95 |
| Leaf Spot(2) | 0.99 | 0.90 | 0.95 |
| **Average** | **0.97** | **0.94** | **0.95** |

The suggested model's class-wise performance shows excellent outcomes in every category shown in TABLE III. With an F1 score of 0.96, Healthy Leaf obtained perfect recall (1.00), meaning all positive samples were accurately identified. The proposed model's balanced and efficient performance was demonstrated by its average precision of 0.97, recall of 0.94, and F1 score of 0.95.

The multi-class ROC curve for our model is shown in Fig. 5. With Healthy_Leaf and Leaf_Rot attaining an AUC of 0.99 and Leaf_Spot earning a perfect AUC of 1.00, all classes perform exceptionally well, demonstrating strong discriminative ability in every category.

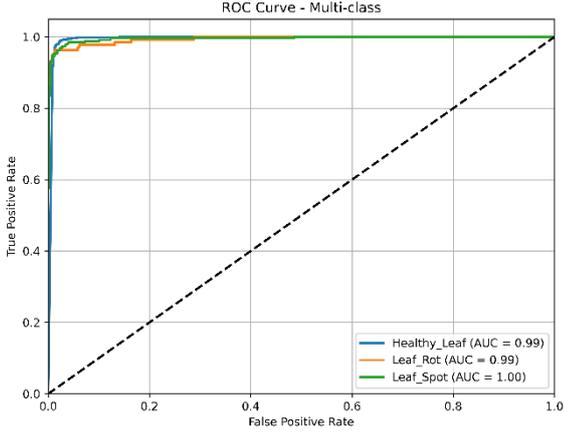

Fig. 5. ROC-AUC curve of the proposed model

The performance of several models, including DenseNet201, VGG16, ResNet50, and the suggested model, is contrasted in TABLE IV. With the highest precision (97%), recall (94%), F1 score (95%), and testing accuracy (95.58%), the suggested model performs better than any other.

TABLE IV. OVERALL PERFORMANCE COMPARISON BETWEEN ALL MODELS

| Model Name | Precision (%) | Recall (%) | F1 Score (%) | Testing Accuracy (%) |
|---|---|---|---|---|
| DenseNet201 | 90 | 90 | 90 | 91.95 |
| VGG16 | 87 | 84 | 86 | 88.42 |
| ResNet50 | 93 | 91 | 92 | 93.52 |
| **Proposed Model** | **97** | **94** | **95** | **95.58** |

With just 2.13 million parameters (~8.13 MB), the suggested model is not only much lighter but also performs better. Comparatively speaking, pre-trained models such as DenseNet201 (~77.6 MB), VGG16 (~528 MB), and ResNet50 (~98 MB) are significantly larger, which makes our model more effective and appropriate for deployment in contexts with limited resources without sacrificing accuracy.

### VI. COMPARISON WITH PREVIOUS WORK

The suggested CBAM-CNN model is compared with other models of previous studies in TABLE V.

TABLE V. RESULTS OF PREVIOUS STUDIES COMPARED WITH THE PROPOSED MODEL

| Reference | Model | Accuracy | XAI |
|---|---|---|---|
| [8] | DenseNet201 | 99.23% | Yes |
| [10] | InceptionV3 | 94.83% | No |
| [16] | SVM-GMM | 83.69% | No |
| [7] | DenseNet201 | 98.77% | No |
| [11] | GLCM-ELM | 97% | No |
| **Proposed Model** | **CBAM-CNN** | **95.58%** | **Yes** |

. A separate dataset different from ours was used to get the 99.23% accuracy achieved by the DenseNet201 model reported in [8]. Comparing performance directly is also a little deceptive because the research in TABLE V all used different datasets. In spite of this, our suggested CBAM-CNN model offers further benefits and achieves a competitive accuracy of 95.58%. The suggested CBAM-CNN is a lightweight, custom-designed model that is more useful for real-time or resource-constrained applications than DenseNet201, which relies on a massive pre-trained architecture with substantial computational and memory overhead. Additionally, our model incorporates explainable AI (XAI) assistance through CBAM in a novel way, offering interpretability that is lacking in the majority of evaluated approaches. When compared to the most recent works, the suggested approach's practical significance is highlighted by the harmony of accuracy, efficiency, and transparency.

### VII. GRAD CAM VISUALIZATION

The model's choice is visually explained via Grad-CAM, which creates a colour-coded heatmap that identifies the areas of a picture that have the greatest influence. Fig. 6 presents heatmap, GradCAM visualization on the same image of a specific class.

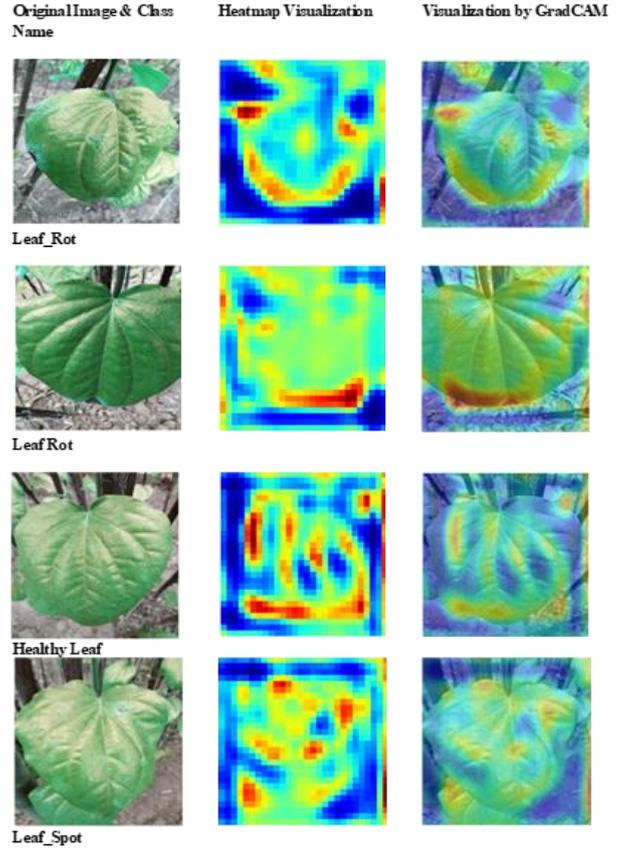

Fig. 6. Heatmap, GradCAM visualization on the same image of a specific class

In this work, the model's predictions for the classification of betel leaf disease were interpreted using Grad-CAM overlays. In order to differentiate between the Healthy Leaf, Leaf Rot, and Leaf Spot classes, the model is focusing on areas of greater importance, which are indicated by warmer colours like red and yellow. These illustrations demonstrate how the model supports its classification judgements by addressing disease-affected areas like lesions, spots, or discolouration patterns. Grad-CAM contributes to explainable AI in agricultural disease detection by localizing attention, which improves interpretability and guarantees that the model's predictions match visual symptoms.

### VIII. LIMITATION AND FUTURE WORK

The current scope of this work is restricted to treating bacterial infections in betel leaf plants, which limits its generalizability to other plant species and disease kinds. Future studies will concentrate on extending the model to other economically significant crops and covering a wider variety of betel leaf illnesses, such as fungal and viral infections. The integration of temporal or time-series data to monitor disease progression and offer practical insights for proactive crop management will also be investigated in future research. The model's resilience in practical settings can be increased by supplementing the dataset with more varied and unbalanced disease classifications. Additionally, to improve classification accuracy with less labelled data, the application of sophisticated semi-supervised and ensemble learning techniques will be examined. Last but not least, creating an intuitive, adaptable interface that offers tailored disease control advice will encourage farmers to implement it in practice, promoting sustainable farming methods and better livelihoods.

IX. CONCLUSION

In conclusion, bacterial infections are the primary cause of betel leaf illnesses, which continuously threaten crop yields worldwide. Deep learning models also play a complete function in farming sectors for crop production growth, cost management, and prediction improvements. This work tackles the urgent problem of early disease identification to protect crop output and food security while highlighting the vital economic and agricultural significance of betel leaf. Through an integrated attention mechanism that adaptively focuses on critical spatial and channel-wise data, we improved disease classification performance by creating a lightweight CBAM-CNN model with just 2.13 million parameters. With 95.58% accuracy, 94% recall, 95% F1 score, and 97% precision, the model outperformed conventional pre-trained CNN architectures while retaining computational economy. Additionally, by emphasizing important disease-relevant areas in the leaves, the use of Grad-CAM as an explainable AI tool confirmed the interpretability of the model. All things considered, this work provides a useful, understandable, and economical method for detecting betel leaf disease, enhancing agricultural output and farmers' incomes.

REFERENCES

[1] "Major Diseases of Betelvine and Their Management," *Crop Diseases and Their Management*, pp. 325–350, Apr. 2016, doi: 10.1201/B19891-24.

[2] "Betel leaf cultivation creeping up as exports expand | The Daily Star." Accessed: Jul. 30, 2025. [Online]. Available: https://www.thedailystar.net/business/economy/news/betel-leaf-cultivation-creeping-exports-expand-3200311

[3] "Finance News: Latest Financial News, Finance News today in Bangladesh." Accessed: Jul. 30, 2025. [Online]. Available: https://today.thefinancialexpress.com.bd/country/annual-turnover-of-rajshahi-betel-leaf-hits-tk-25-billion-1708013923

[4] "Betel leaf cultivation increasing in Barguna." Accessed: Jul. 30, 2025.[Online].Available:https://www.agri24.tv/english/arable/news/598

[5] P. Biswas et al., "Betelvine (Piper betle L.): A comprehensive insight into its ethnopharmacology, phytochemistry, and pharmacological, biomedical and therapeutic attributes," *Journal of Cellular and Molecular Medicine*, vol. 26, no. 11, pp. 3083–3119,Jun.2022,doi:10.1111/JCMM.17323;JOURNAL:JOURNAL:15824934;PAGE:STRING:ARTICLE/CHAPTER.

[6] Z. EL Housni, T. Abdessalem, N. Radouane, S. Ezrari, A. Zegoumou, and A. Ouijja, "Overview of sugar beet leaf spot disease caused by Cercospora beticola Sacc," *Archives of Phytopathology and Plant Protection*, vol. 56, no. 7, pp. 503–528, 2023, doi: 10.1080/03235408.2023.2216356.

[7] S. Kusuma and K. R. Jothi, "Early betel leaf disease detection using vision transformer and deep learning algorithms," *International Journal of Information Technology (Singapore)*, vol. 16, no. 1, pp. 169–180, Jan. 2024, doi: 10.1007/S41870-023-01647-3/METRICS.

[8] M. Tahsin et al., "Leveraging pre-trained models within a semi-supervised and explainable AI RealTime framework: A pioneering paradigm for betel leaf disease detection," *Journal of Agriculture and Food Research*, vol. 22, p. 102142, Aug. 2025, doi: 10.1016/J.JAFR.2025.102142.

[9] R. H. Hridoy, M. Tarek Habib, M. Sadekur Rahman, and M. S. Uddin, "Deep Neural Networks-Based Recognition of Betel Plant Diseases by Leaf Image Classification," *Lecture Notes on Data Engineering and Communications Technologies*, vol. 116, pp. 227–241, 2022, doi: 10.1007/978-981-16-9605-3_16.

[10] M. A. Malek, S. Basak, and S. S. Reya, "An Approach to Identify Diseases in Betel Leaf Using Deep Learning Techniques," *2022 4th International Conference on Sustainable Technologies for Industry 4.0, STI 2022*, 2022, doi: 10.1109/STI56238.2022.10103348.

[11] F. David and M. A. Mukunthan, "Betel Leaf Diseases Classification using Machine Learning Algorithm: A Feasible Approach," *Journal of Advanced Research in Applied Sciences and Engineering Technology Journal homepage*, vol. 40, pp. 74–86, 2024, doi: 10.37934/araset.40.1.7486.

[12] H. Hamdani, A. Septiarini, N. Puspitasari, A. Tejawati, and F. Alameka, "The color features and k-nearest neighbor algorithm for classifying betel leaf image," *IAES International Journal of Robotics and Automation (IJRA)*, vol. 13, no. 3, pp. 330–337, 2024, doi: 10.11591/ijra.v13i3.pp330-337.

[13] M. Gayakwad et al., "Applying the transfer learning models on the dataset on the effect of diseases on Nagvel-betel (Piper betle) leaves," *Data in Brief*, vol. 62, p. 111987, Oct. 2025, doi: 10.1016/J.DIB.2025.111987.

[14] R. H. Hridoy, M. T. Habib, I. Mahmud, A. Haque, and M. A. Al Mamun, "A comprehensive image dataset for accurate diagnosis of betel leaf diseases using artificial intelligence in plant pathology," *Data in Brief*, vol. 60, p. 111564, Jun. 2025, doi: 10.1016/J.DIB.2025.111564.

[15] S. Woo, J. Park, J.-Y. Lee, and I. S. Kweon, "CBAM: Convolutional Block Attention Module".

[16] M. Z. Hasan, N. Zeba, A. Malek, and S. S. Reya, "A Leaf Disease Classification Model in Betel Vine Using Machine Learning Techniques," *International Conference on Robotics, Electrical and Signal Processing Techniques*, pp. 362–366, 2021, doi: 10.1109/ICREST51555.2021.9331142.